\documentclass[10pt,twocolumn,letterpaper]{article}

\usepackage[pagenumbers]{cvpr} 
\usepackage{algorithm}
\usepackage{algpseudocode}
\usepackage{amsmath}
\usepackage{placeins}

\usepackage{float}

\algnewcommand\Optional{\item[\textbf{Optional:}]}
\usepackage{physics}
\usepackage{makecell}
\usepackage{booktabs}

\definecolor{cvprblue}{rgb}{0.21,0.49,0.74}
\usepackage[pagebackref,breaklinks,colorlinks,allcolors=cvprblue]{hyperref}

\title{Block Cascading: Training Free Acceleration of Block-Causal Video Models\vspace{-0.5cm}}

\author{
Hmrishav Bandyopadhyay$^{1,2}$ \quad
Nikhil Pinnaparaju$^{1}$ \quad
Rahim Entezari$^{1}$ \\[2pt]
Jim Scott$^{1}$ \quad
Yi-Zhe Song$^{2}$ \quad
Varun Jampani$^{1}$ \\[6pt]
$^{1}$Stability AI \quad
$^{2}$SketchX, University of Surrey
}

\begin{document}

\twocolumn[{%
\renewcommand\twocolumn[1][]{#1}%
\maketitle
    \captionsetup{type=figure}
    \vspace{-1cm}
    \includegraphics[trim={0 0 1cm 0}, width=\textwidth]{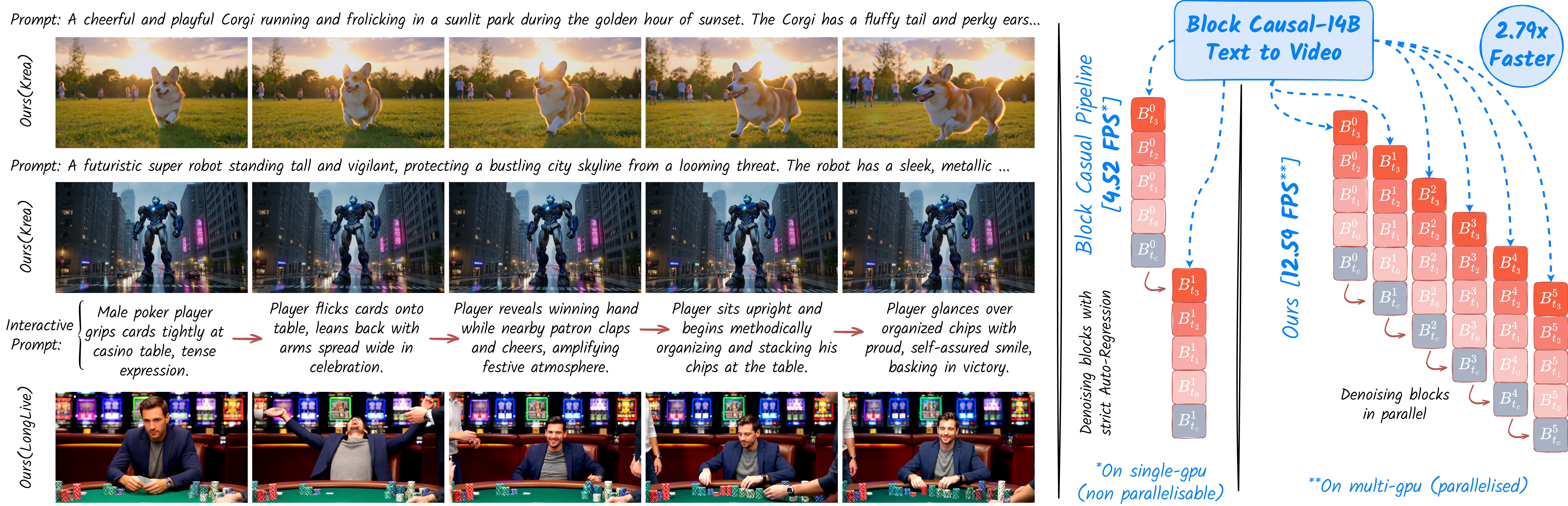} \\[-0.6cm]   
    \caption{ We propose Block Cascading (right), where future blocks of frames are cascaded with current ones to improve generation speed. Without requiring any fine-tuning, Block Cascading can improve inference FPS by upto 2.79x (right) in block-causal pipelines without negatively affecting quality for short, long, and interactive video generation (left).
    \textbf{All FPS is reported using the same python environment with Flash Attention 3~\cite{shah2024flashattention} on H100 GPUs. Most prompts are condensed in all figures, with full prompts available in Supplementary PDF and all videos available in Supplementary zip}.}
    \label{fig:teaser}
    \vspace{0.2cm}
}]


\begin{abstract}
Block-causal video generation faces a stark speed-quality trade-off: small 1.3B models manage only 16 FPS while large 14B models crawl at 4.5 FPS, forcing users to choose between responsiveness and quality. Block Cascading significantly mitigates this trade-off through training-free parallelization. Our key insight: future video blocks do not need fully denoised current blocks to begin generation. By starting block generation with partially denoised context from predecessors, we transform sequential pipelines into parallel cascades where multiple blocks denoise simultaneously. With 5 GPUs exploiting temporal parallelism, we achieve $\sim$$2$× acceleration across all model scales: 1.3B models accelerate from 16 to 30 FPS, 14B models from 4.5 to 12.5 FPS. Beyond inference speed, Block Cascading eliminates overhead from KV-recaching (of $\sim$$200$ms) during context switches for interactive generation. Extensive evaluations validated against multiple block-causal pipelines demonstrate no significant loss in generation quality when switching from block-causal to Block Cascading pipelines for inference.
\end{abstract}

\vspace{-0.4cm}
\section{Introduction}

Diffusion-based block-causal video generation pipelines denoise videos auto-regressively for faster, streamable generation. However, generation speed decreases significantly with model scale~\cite{wan2025wan, kong2024hunyuanvideo}. Small-scale diffusion models (1.3B parameters) barely reach 16 FPS~\cite{huang2025self}, while high-quality large-scale models (14B parameters~\cite{krea_realtime_14b}) crawl at 4.5 FPS on H100 GPUs, making them impractical for any interactive use. This speed-quality trade-off forces users to choose between responsive but lower-quality small models, or impressive but unusably slow large models. We revisit this trade-off, arguing that beyond model complexity it stems from rigid scheduling with strict causal dependencies. To overcome this, we introduce parallelised video generation with Block Cascading, unlocking faster inference with multiple GPUs and smarter scheduling.

The key intuition driving our work is that future frames need not rely on fully denoised past frames to begin generation. Current block-causal models like CausVid~\cite{causvid}, Self-Forcing~\cite{huang2025self}, and LongLive~\cite{yang2025longlive} enforce strict sequential dependencies: block $B^{i+1}$ must wait for block $B^i$ to completely denoise from $t{=}1000$ to $t{=}0$ before starting. This rigid ordering allows the denoising of only one block at a time in a causal manner, limiting opportunities to parallelise generation with multiple GPUs. We demonstrate that this strict dependency is overly conservative, especially since many of these block-causal models are adapted from bidirectional pre-trained networks~\cite{wan2025wan} that naturally accommodate noisy context. Empirically, we observe that conditioning $B^{i+1}$ on partially denoised outputs from $B^{i}$ at intermediate timesteps (\eg $t{=}750$) can yield visual quality comparable to using fully denoised blocks.

To build on this, we propose Block Cascading to overlap denoising of multiple blocks for enabling parallel generation across GPUs. Instead of sequential processing ($B^1$ completes, then $B^2$ starts, then $B^3$ starts), we cascade block generation: when $B^1$ reaches an intermediate timestep, we use it's features as noisy context to start denoising $B^2$. Similarly, $B^3$ begins denoising once $B^2$ reaches this intermediate checkpoint. This transforms the generation pipeline from sequential to parallel where multiple blocks denoise simultaneously at different stages, significantly reducing inference time. In effect, this reduces the waiting time for future blocks, which are already partially generated when the current block is decoded and viewed. Further, by cascading current and future blocks together, we can leverage bidirectional attention amongst these blocks to potentially improve generation quality.

Pre-generation of future blocks helps exploit their noisy context during prompt switches for interactive video generation. In block-causal pipelines, context switches have to be accompanied by expensive KV-recaching, where past KV is re-cached using new prompts, causing 200ms+ latency spikes which break interactivity. With Block Cascading, we instead switch context by simply changing the text prompt as future blocks demonstrate a gradual adaptation to new context based on their current noise level. This allows for seamless context switching without blocky artifacts or abrupt changes in scenes and subjects.

Finally, we note that our multi-GPU pipeline exploits temporal parallelism: each GPU processes a different temporal segment of the video. This approach is fundamentally distinct from data parallelism (processing independent samples) or model parallelism (splitting layers). However, sub-linear scaling (2.79× FPS on 5 GPUs with a 14B model) is expected due to VAE decoding after generation and overhead from KV communication across GPUs for self-attention. Optimizations targeting these factors, such as linear attention, quantization, or smaller VAEs can be applied on top of block-cascaded models for additional gains. Moreover, VAE decoding can be moved to a separate GPU to overlap decoding with next-block denoising. We do not discuss these optimizations in detail as they are not unique to block cascading.

Our contributions are: \textit{(i)} We identify that strict causal dependencies in video generation are unnecessary, as noisy intermediate states provide sufficient context for generation; \textit{(ii)} We propose Block Cascading, a training-free parallelization strategy that works universally across different block-causal models; \textit{(iii)} We achieve practical inference speed with 5 GPUs: 30 FPS for 1.3B models and 12.5 FPS for 14B models, while maintaining quality comparable to baselines; \textit{(iv)} We demonstrate prompt switching without latency spikes in interactive applications, previously impossible with causal models. Across short~\cite{huang2025self}, long~\cite{yang2025longlive}, and interactive~\cite{yang2025longlive} video generation, and even across model scales~\cite{krea_realtime_14b}, 
Block Cascading consistently delivers significant speed-up (average $\sim$$2$× with 5 GPUs, 10\% on single GPU) without any fine-tuning.

\vspace{-0.2cm}
\section{Related works}
\label{sec:related}
\vspace{-0.1cm}
\subsection{Timestep Distillation}

Timestep Distillation~\cite{zheng2024trajectory, yin2024one, yin2024improved, chen2024nitrofusion, lin2024sdxl} of diffusion models~\cite{ho2020denoising} compresses their diffusion ODE trajectory~\cite{song2020score} for speeding up iterative denoising from noise to data. This setup typically involves using a multi-step teacher model to train a few-step student by comparing their outputs along different points of the trajectory. Progressive distillation~\cite{salimans2022progressive, lin2024sdxl} directly compares teacher and student outputs for same inputs in short skips of ODE trajectory. Advanced approaches like DMD~\cite{yin2024one, yin2024improved} and SiD~\cite{zhou2024score} use scores from teacher and student models to align the distributions of multi-step teacher and few-step student. Recent works can perform aggressive ODE compression~\cite{chen2025sana-sprint}, making it possible to generate in as few as one step~\cite{yin2024one, chen2024nitrofusion}, generally by supplementing distribution matching with adversarial techniques~\cite{lin2024sdxl, chen2024nitrofusion} and progressive distillation~\cite{yin2024one} ideas. We perform few-step inference with timestep-distilled models, to reduce the complexity of Block Cascading inference.

\vspace{-0.1cm}
\subsection{Auto-Regressive Video Generation} Early Autoregressive models~\cite{yan2021videogpt, tian2021good} have been predominantly used in conjunction with adversarial training~\cite{tian2021good, skorokhodov2022stylegan, wang2023styleinv} for sequential frame-by-frame video generation. With the advancement of diffusion-based generative models for image generation~\cite{esser2024scaling, ho2020denoising, rombach2022high}, recent works in video generation~\cite{wang2023modelscope,yang2024cogvideox,wan2025wan,ma2025stepvideot2vtechnicalreportpractice,kong2024hunyuanvideo, chen2025sana} have adopted diffusion models to generate all video frames at once with full bidirectional attention across frames. These bidirectional models~\cite{wan2025wan, yang2024cogvideox, kong2024hunyuanvideo} can generate high quality videos but are slow and bounded by quadratic attention scaling~\cite{wang2020linformer} with respect to number of generated frames. 

To improve speed in video generation pipelines, recent works~\cite{zhang2024sf,lin2025diffusion, mao2025osv, zhang2025fast,zhang2025vsa} attempt to reduce number of sampling steps to directly reduce generation latency through timestep distillation. In parallel, several works re-introduce auto-regressive generative modelling of videos~\cite{weng2024art, gao2024vid,causvid, huang2025self, liu2025rolling, yang2025longlive, lin2025autoregressive} by distilling bidirectional models~\cite{causvid, huang2025self} or training auto-regressive pipelines from scratch~\cite{teng2025magi, chen2025skyreels}. Fast, causal video generative pipelines set the stage for interactive and controllable generation~\cite{yang2025longlive, yu2025gamefactory, shin2025motionstream}, letting users inject control and action prompts that can influence future frames. In this work, we relax causality restrictions in distilled auto-regressive models to speed up inference by parallel processing of future blocks.

\vspace{-0.1cm}
\subsection{KV Caching} 
Autoregressive generation of videos~\cite{yang2025longlive, huang2025self} can be accelerated during inference with KV caching, where Key-Value (KV) pairs of previous frames are retained for efficient attention computation~\cite{vaswani2017attention}. Since information from previous frames provides auto-regressive models with context for current frame generation, KV caching~\cite{radford2019language} helps prevent recomputing these features by simply storing them. KV caching was originally introduced in LLMs~\cite{vaswani2017attention,dai2019transformer, brown2020language, xiao2023efficient, liu2023ring,hooper2024kvquant} and has been since adapted to video generation~\cite{gao2024vid, causvid, huang2025self, yang2025longlive} to improve generation efficiency. While KV caching reduces KV computation for attention, the attention unit itself scales at a quadratic rate~\cite{wang2020linformer} with KV size, requiring eviction of older cache elements time to time to support longer generation. Evicting salient features, however, can cause issues like drifting~\cite{zhang2025framepack} where the model forgets key context from previous time-frames. To reduce attention complexity without drifting, both LLMs~\cite{xiao2023efficient, su2025kvsink} and video generation approaches~\cite{huang2025self, yang2025longlive, liu2025rolling} use some form of KV sink~\cite{su2025kvsink}, where key features from earlier frames are retained. In this paper, we share KV features from parallel blocks to compute self-attention, reducing the requirement of an external KV cache during inference.

\vspace{-0.2cm}
\section{Methodology}
\begin{figure*}
    \centering
    \vspace{-0.1cm}
    \includegraphics[width=0.97\linewidth]{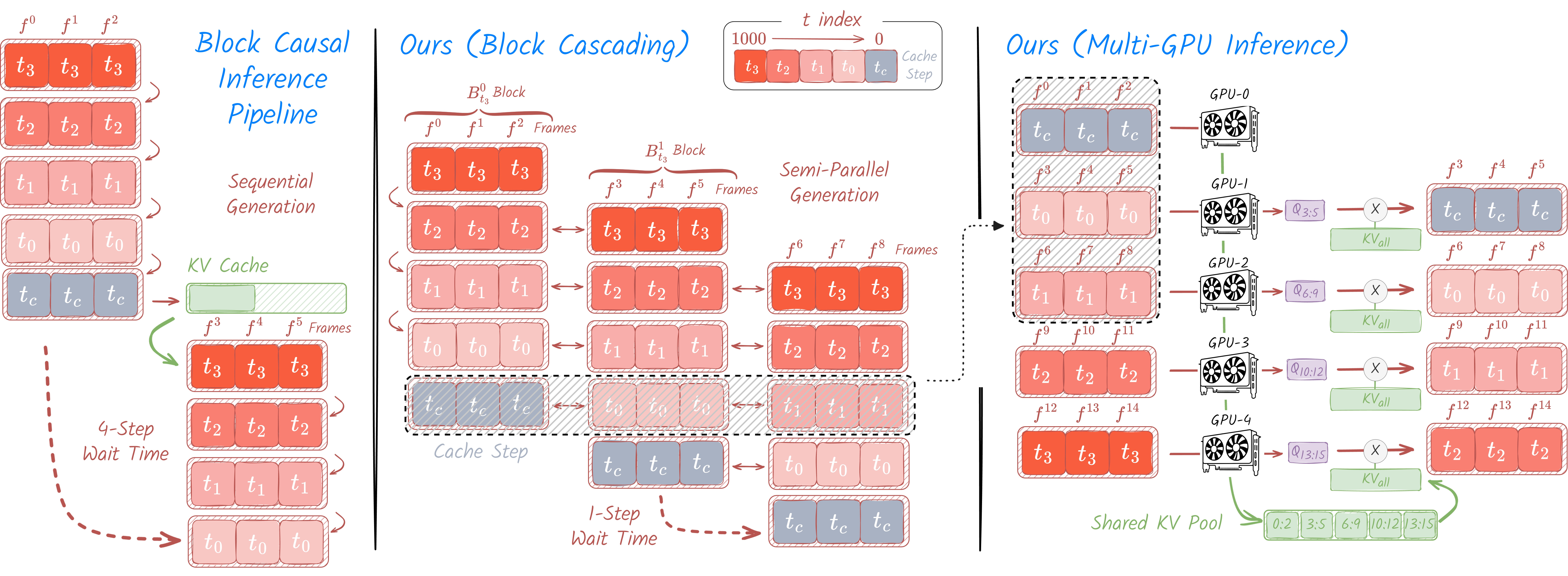}
    \vspace{-0.3cm}
    \caption{\textbf{Block Cascading Inference:} Our inference pipeline reduces dependency on clean, denoised blocks for future block generation. For example in here (center), we denoise block $B^1_{t_3}{=}\{f^3,f^4, f^5\}$ and $B^0_{t_2} {=}\{f^3,f^4, f^5\}$ jointly using bidirectional attention instead of waiting for $B^0_{t_c}$ to denoise  $B^1_{t_3}$ (left)~\cite{causvid, huang2025self, yang2025longlive}. By reducing dependency on previous blocks, we can free up the inference pipeline and allow parallel processing of multiple blocks (right) to improve generation speed significantly. }
    \label{fig:main_diag}
    \vspace{-0.6cm}
\end{figure*}

\label{sec: method}
\subsection{Background and Overview}
\label{sec: overview}

Diffusion based multi-step generative models~\cite{wan2025wan, kong2024hunyuanvideo, yang2024cogvideox} have predominantly used bidirectional attention for high quality video generation at the cost of extra compute and generation time. Recent few-step video generation works like CausVid~\cite{causvid} and Self-Forcing~\cite{huang2025self} reduce generation time by distilling multi-step to few-step diffusion and by modelling auto-regressive generation on top of bidirectional models \cite{wan2025wan}. 

Adapting bidirectional pipelines like Wan2.1\cite{wan2025wan} for auto-regressive causal generation is non-trivial as it requires training to accommodate for the fundamental distribution shift where current frames $x^i$ are modelled through previous frames $x^{<i}$ only. This is usually in the form of a regression-based objective including points on the teacher and student trajectories where the student is a causal network. This student is further trained as a few-step model with a timestep-distillation objective to catch up to the multi-step bidirectional teacher. CausVid trains from Wan2.1~\cite{wan2025wan} by matching distributions of the Wan2.1 teacher ($p_\text{real}$) and the causal few-step student ($p_\text{fake}$) as:  

\vspace{-0.4cm}
\begin{equation}
D_{\text{KL}} (p_\text{fake} || p_\text{real}) = - \mathbb{E}_{x \sim p_\text{fake}}\Big(\log \frac{p_\text{real}(x)}{p_\text{fake}(x)}\Big)
\label{eq:first_kl}
\end{equation}
for samples $x$ generated by the student. This KL divergence being intractable, the DMD objective allows representing the gradient of KL as a difference of scores~\cite{song2020score, zhou2024score,wang2023prolificdreamer, poole2022dreamfusion} from the teacher and the student model as: 

\vspace{-0.4cm}
\begin{equation}\label{eq:kldiv}
    \nabla_\theta \mathcal{L}_\text{DMD} = - \mathbb{E}_{x \sim p_\text{fake}} \Big((s_\text{real}(x) - s_\text{fake}(x))\dv{G_\theta}{\theta}\Big) 
\end{equation}

\noindent where $G_\theta$ corresponds to the student network, and $s_\text{fake}$ and $s_\text{real}$ are the teacher and student scoring functions respectively. In practice, we use the teacher network for $s_\text{real}$ and a proxy $s_\text{fake}$ trained to adapt to the changing distribution of the student model $G_\theta$. To provide the proxy model enough room for catching up to the generator, it is trained for $N$ iterations at every generator update.

\noindent \textbf{Block-causal video generation:} To balance quality and speed, CausVid~\cite{causvid}, Self-Forcing~\cite{huang2025self} and other recent works~\cite{yang2025longlive, liu2025rolling} employ a block-causal (rather than strictly causal) pipeline. In here, latent frames are temporally bunched into blocks $B^i_t{=}\{f^k_{t_n} : k \in K\}$, which are processed in a causal (\ie $B^0 \to B^1 \to B^2 \ldots$) fashion. Internally each block with it's group of frames uses bidirectional attention to generate video segments. During inference, blocks are rolled out sequentially, and denoised from $t{=}1000$ to $t{=}0$.

CausVid~\cite{causvid} and Self-Forcing~\cite{huang2025self} introduce efficient causal video generation through KV caching~\cite{pope2023efficiently}, where future blocks can depend on pre-computed Key-Value (KV) features of previous blocks during self-attention. This prevents recomputing KV features for obtaining context from previous blocks which have been denoised. To generate long videos and improve interactive control in generated frames, LongLive~\cite{yang2025longlive} introduces KV-recaching where stored KV is recalculated for new prompts. This form of KV caching where future blocks depend on clean KV from previous blocks requires future blocks to wait for current block to finish denoising. Further in the event of a KV-recache, all previous cache has to be discarded and recomputed - causing a drop in inference FPS.

In this work, we propose to remove this \textit{stall} by denoising noisy future blocks $B^{i+1}_{t}$ with noisy context in current block $B^{i}_{t-1}$. In other words, we \emph{speculatively} roll forward with a \textit{noisy cache} consisting of partially denoised KV features to get started with video generation in future blocks.

\subsection{Noisy Caching}

For Block-wise Causal generation of videos, blocks $B_t^i{=}\{f^j_t, f^{j+1}_t, f^{j+2}_t\}$ with frames $f^*$ and $t>0$ are denoised using previously denoised blocks $B^{<i}_0$. In a multi-step diffusion environment, the block $B^i$ has to wait till all previous blocks $B^{<i}_0$ are denoised to $t{=}0$ for using their \textit{clean} KV features in block-causal self-attention. However, block-causal distilled models \cite{causvid, huang2025self, yang2025longlive} are often trained from bidirectional teachers~\cite{wan2025wan} that use noisy features from all frames with bidirectional attention. This allows us to exploit the underlying bidirectional pre-training of causal distilled models to kickstart frame generation in future blocks even when current block is not completely denoised.

Specifically, we start denoising block $B^i_{T}$ at $t{=}T$ using noisy cache from blocks $\{B^{i-k}_{T-k}\}_{k=1}^{i}$ instead of waiting for blocks $B^{<i}$ to finish denoising. Again, at $t{=}t_0$, the previous blocks $B^{<i}$ have already finished denoising, but we retain context for $B^i_{t_0}$, through another forward pass, using $B^{i-1}_{t_c}$ where $t_c{=}0$ corresponds to the cache timestep. This is similar to how previous works~\cite{huang2025self, causvid} construct KV cache from denoised blocks with an extra forward pass. Forward pass for block $B^i_{t_0}$ thus requires attending to KV from $\{B^{i-k}_{t_c}\}_{k=1}^{i}$ in addition to it's own for self-attention.

In practice however, it is not computationally feasible to record noisy KV for all previous blocks $B^{<i}$ on a $N{=}50$ step diffusion model like Wan2.1\cite{wan2025wan}, even if just for inference. Working with a limited window size of $M$ blocks, this has a complexity of $O(N \times M \times M)$, with Dot-Product Attention complexity of $O(M)$ for KV size of $M$ and a fixed number of frames per block. Since complexity linearly scales with number of steps, we can directly reduce the complexity of this operation using distillation techniques~\cite{yin2024one, yin2024improved, chen2024nitrofusion} that reduce inference timesteps to improve feasibility. We use $4$ steps to denoise from $T{=}1000$ (i.e. $\{t_3,t_2,t_1,t_0\}$) and an additional step $t_c{=}0$ to improve video consistency across different blocks with cleaner cache from previous frames.

Naive implementation of this 4 step semi-causal denoising brings us to another sequential trap where we denoise the current block $B^{i}_t \to B^{i}_{t-1}$ and then start denoising the next block $B^{i+1}_{t}$ using KV from $B^i_{t-1}$ sequentially. This kickstarts future block denoising but slows down current block, resulting in slow inference. Concurrent work Rolling-Forcing~\cite{liu2025rolling} proposes to stitch the blocks together and jointly denoise all frames, following previous research on robust training for causal video models~\cite{chen2024diffusion}. Fundamentally different to Rolling Forcing~\cite{liu2025rolling} and Diffusion Forcing~\cite{chen2024diffusion}, we propose denoising of each block \textit{individually} and \textit{parallelly} through mini-batching, following individual block denoising in CausVid~\cite{causvid} and Self-Forcing~\cite{huang2025self}. We construct mini-batches where previous and current blocks are batched together with shared attention. (see~\cref{fig:main_diag}).

\vspace{-0.2cm}
\subsection{Block Cascading\label{sec:blockcascading} }
\vspace{-0.1cm}
Mini-batching multiple blocks together can overburden a single GPU and reduce generation FPS (see~\cref{fig:abl-single}). To reduce the computation burden of our pipeline on a single GPU, we process batches on different GPUs and share KV features for computing self-attention. Using a fixed window size of $M$, we batch together blocks that depend on each other for denoising and can share KV during denoising. For denoising each mini-batch, we share KV features of each block with other blocks being denoised at each self attention layer. This is done by creating a global \textit{shared KV pool} which can be referred to by all blocks, enabling continuity in generated frames. 
By nature, this KV sharing mimics rolling KV cache in block-causal generation pipelines ~\cite{huang2025self, causvid} that allows generation of long videos without running out of memory with fixed window sizes. Maximum parallelisation can be attained with a window-size equal to number of denoising steps ( with $t_c$) where each block depends on previous block's cleaner timestep. Increasing window size beyond this requires maintaining an add on KV cache from previous blocks (similar to~\cite{huang2025self}). While smaller window sizes are possible by restricting attention window during self-attention, we find they can impact generation as later blocks do not get access to clean latents at $t{=}t_c$. A better way to obtain smaller windows is through reduced parallelism (see ~\cref{sec: abl}).

We note that generating long videos with a fixed window often results in drifting artifacts~\cite{zhang2025framepack}. Following recent works ~\cite{yang2025longlive}, we reduce drifting in long video generation by using the first block as a `sink' token, that stays in an external KV cache. 

{
\setlength{\textfloatsep}{0pt}
\begin{algorithm}
\footnotesize
\caption{Multi-GPU Inference with Block Cascading}
\label{alg:ar_diffusion_kv}
\begin{algorithmic}[1]
\Require Generator $G_\theta$, Timesteps $\mathcal{T} = \{t_3, t_2, t_1, t_0, t_c\}$
\Require Total Frames $M$, Block Size $S$, GPUs $\{g_1, \ldots, g_G\}$, Scheduler $\psi$, VAE Decoder $\mathcal{D}$
\State \textbf{Block computation:}
\State $B \leftarrow \lceil M / S \rceil$ \Comment{Number of blocks}
\State \textbf{Initialize:}
\State $\mathbf{X} \sim \mathcal{N}(0, I)$ \Comment{Noisy latents, size $M$}
\State $\mathbf{O} \leftarrow \emptyset$ \Comment{Output latents}
\State $b \leftarrow 0$ \Comment{Primary block index}
\State $\tau \leftarrow 0$ \Comment{Timestep index in $\mathcal{T}$}
\State $d \leftarrow 0$ \Comment{Cascade depth}
\While{$d \geq 0$}
    \State $\mathcal{B} \leftarrow \bigcup_{i=0}^{\min(d, G-1)} \{(\mathbf{x}_{b+i}, \mathcal{T}[\tau - i])\}$ on $g_i$ \Comment{Create Batch}
    \State $\{\hat{\mathbf{x}}_0^{g_0}, \ldots, \hat{\mathbf{x}}_d^{g_d}\} \leftarrow G_\theta(\mathcal{B})$
    \For{each $\hat{\mathbf{x}}_j^{g_j}$}
        \If{$\mathcal{T}[\tau - j] > t_0$}
            \State $\epsilon \sim \mathcal{N}(0, I)$
            \State $\mathbf{X}[b+j] \leftarrow \Psi(\hat{\mathbf{x}}_j^{g_j}, \epsilon, \mathcal{T}[\tau - j + 1])$
        \ElsIf{$\mathcal{T}[\tau - j] = t_0$}
            \State $\mathbf{O}[b+j] \leftarrow \hat{\mathbf{x}}_j^{g_j}$ \Comment{For streaming, decode here}
        \EndIf
    \EndFor
    \If{block $b$ completed $t_c$}
        \State $b \leftarrow b + 1$; $d \leftarrow d - 1$ if $b + d \geq B$
    \Else
        \State $d \leftarrow \min(d + 1, G-1)$; $\tau \leftarrow \tau + 1$
    \EndIf
\EndWhile
\State \Return $\mathcal{D}(\mathbf{O})$ \Comment{VAE decode}
\end{algorithmic}
\end{algorithm}
}
\begin{figure}[th]
    \centering
    \vspace{-0.4cm}
    \includegraphics[width=\linewidth]{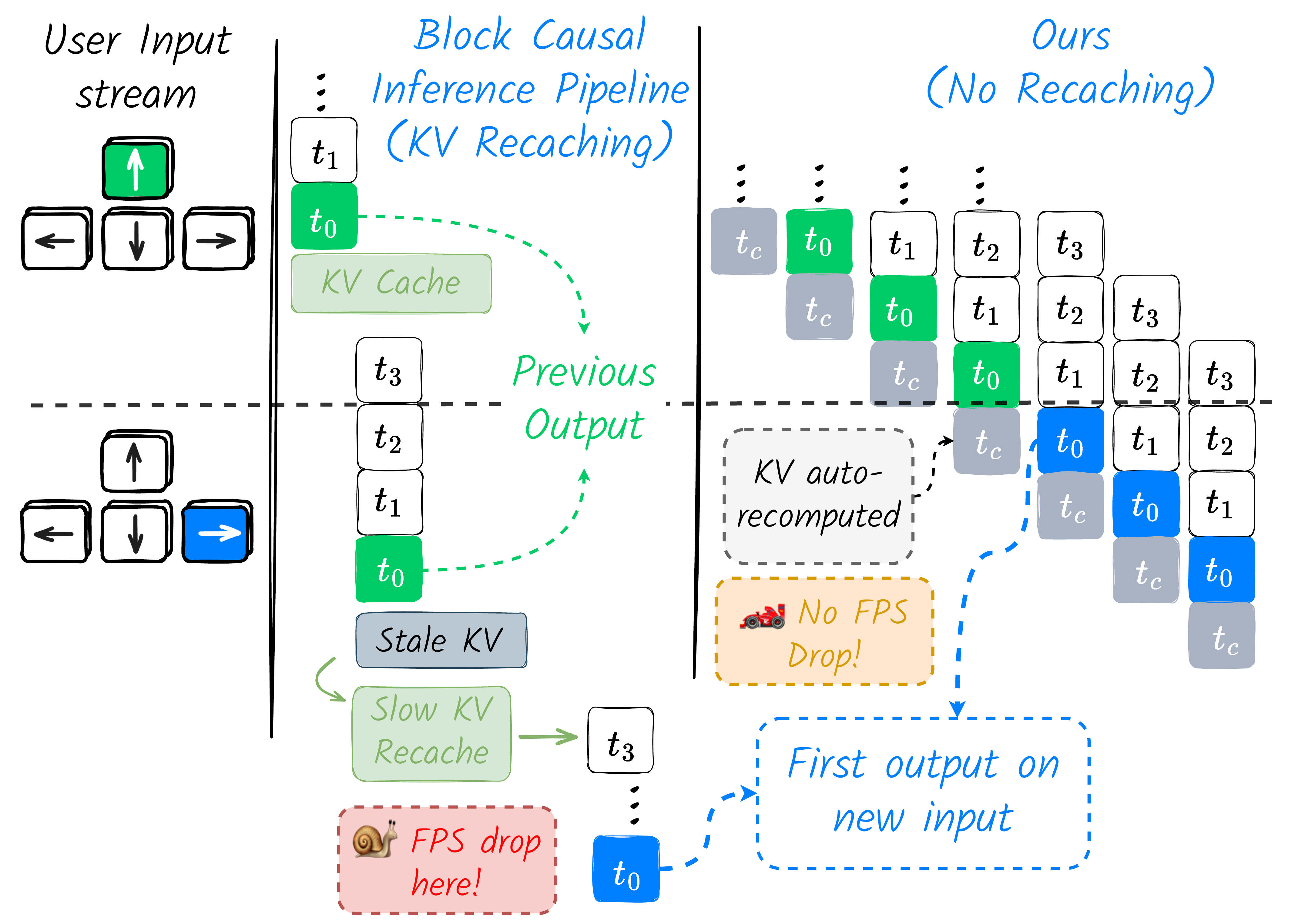}
    \vspace{-0.6cm}
    \caption{\textbf{Expensive KV-recaching: } KV-recaching can result in FPS drop in interactive generation as previous generated and stored cache has to be recached using new context. We skip KV-recaching as our KV is auto-recomputed using new context.}
    \label{fig:recache}
    \vspace{-0.2cm}
\end{figure}

\vspace{-0.2cm}
\subsection{Training Free Adaptation} 

Naive causal denoising with Block Cascading can cause artifacts, specifically in fine-grained features like colour and pattern on a surf-board in~\cref{fig:causal_artifacts}. These artifacts are noticeable in video models as frame-jumps, where fine-grained details are not aligned across frames. We find that artifacts are prominent particularly in early frames where the model does not have enough context from clean latents (\ie $B^{<i}_{t_c}$). For example, during denoising $B^1$ using KV features from $B^0$, $B^1$ observes clean features from $B^{0}_{t_c}$ only during it's last time-step, during denoising $B^{1}_{t_0}$. This creates occasional distortion from context mismatch between previously generated context $B^1_{t_0}$ and new cache context $B^{0}_{t_c}$. Consequently, we find these distortions missing in ablative experiments with lower parallelism (see \cref{sec: abl}) where previously generated context (\eg $B^1_{t_2}$) is still too noisy for conflict. To fix these artifacts and generate smooth videos, we exploit bidirectional pre-training and use full bidirectional attention across the current cascade. Bidirectional attention here helps smooth out inconsistencies by naturally aligning the features of both current and future frames during video generation (see ~\cref{fig:causal_artifacts}). Since bidirectional attention uses more context to denoise every block, it can even surpass the block-causal baseline in quality at times (see ~\cref{fig:user_study}).

\vspace{-0.1cm}
\subsection{Interactive Control}
\begin{figure*}[!h]
    \centering
    \begin{subfigure}[b]{0.33\textwidth}
        \centering
        \includegraphics[width=\textwidth]{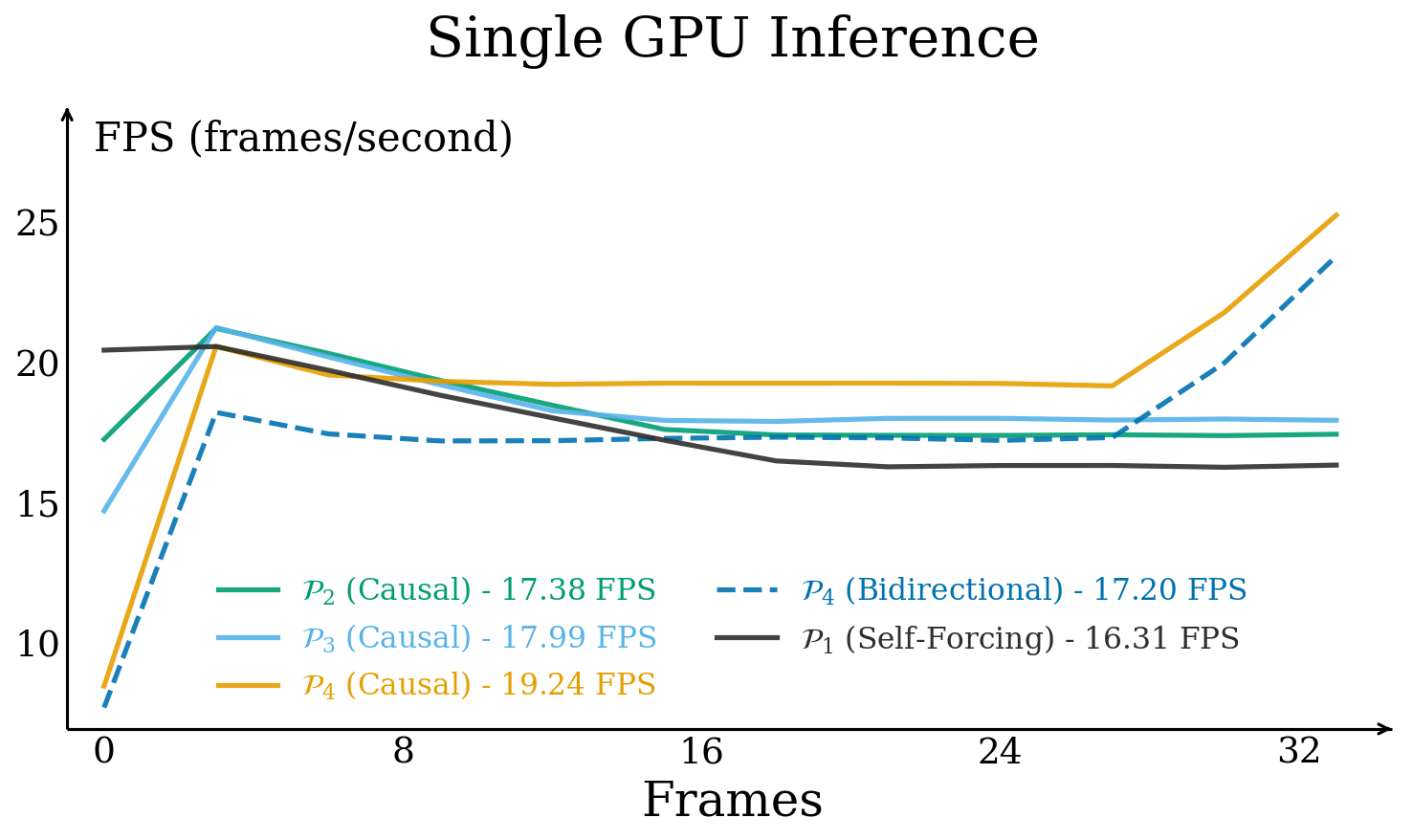}
        \caption{Ablative config. in Single GPU environment}
        \label{fig:abl-single}
    \end{subfigure}
    \hfill
    \begin{subfigure}[b]{0.33\textwidth}
        \centering
        \includegraphics[width=\textwidth]{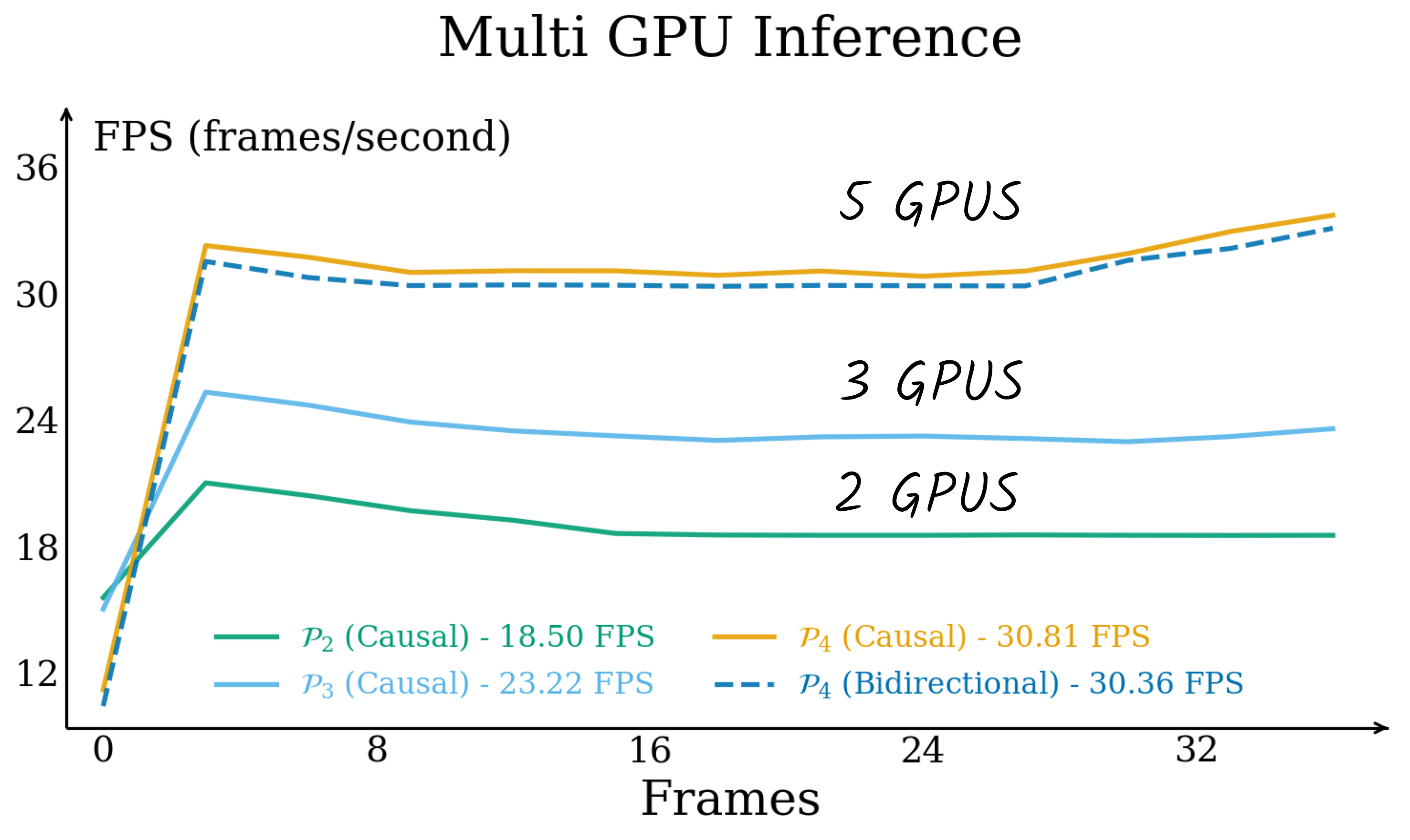}
        \caption{Ablative config. in Multi GPU environment}
        \label{fig:abl-multi}
    \end{subfigure}
    \hfill
    \begin{subfigure}[b]{0.33\textwidth}
        \centering
        \includegraphics[width=\textwidth]{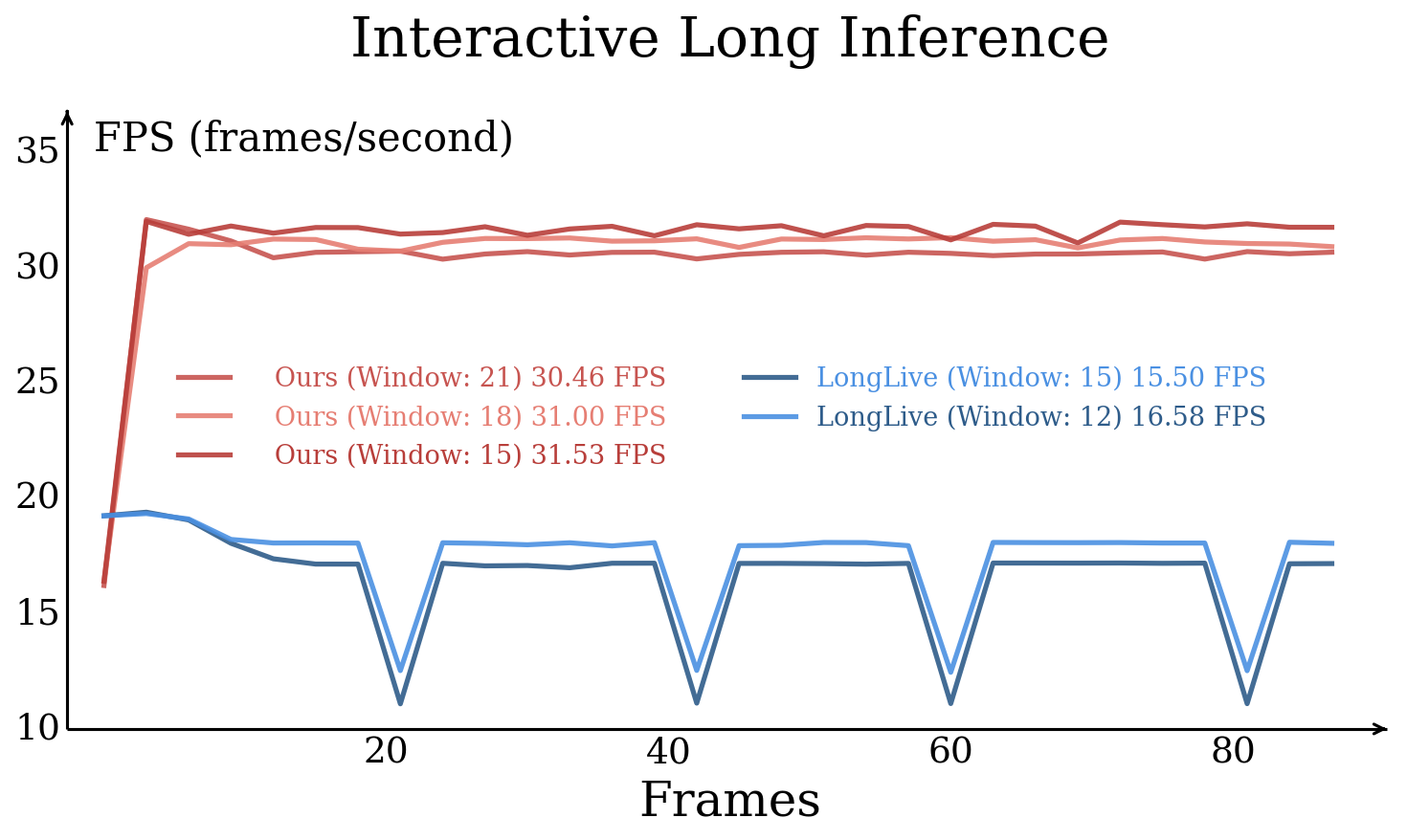}
        \caption{FPS drop from KV-recaching}
        \label{fig:fps_recache}
    \end{subfigure}
    \vspace{-0.6cm}
    \caption{\textbf{Instantaneous FPS:} We measure FPS as time taken to denoise a particular block of 3 latent frames (12 video frames). FPS fluctuates from changing attention window sizes during block-causal denoising and number of parallel blocks in Block Cascading.}
    \label{fig:fps-combined}
    \vspace{-0.6cm}

\end{figure*}

To allow interactive control over video generation, current block-causal denoising pipelines ~\cite{yang2025longlive} perform `KV-recaching', recomputing past KV cache with new prompts for influencing next block generation. While starting denoising at $t{=}1000$, current blocks are heavily influenced by clean past cache, and without KV-recaching, can completely ignore new instructions. However, KV-recaching heavily drops inference FPS as all previous blocks in the attention window have to be recached before current block can start generating (see ~\cref{fig:fps_recache}). We demonstrate that KV-recaching can be avoided with our Block Cascading pipeline since we adapt to changing context easily with new text prompts. Specifically during a context switch, future blocks are at various stages of denoising while the first block is denoised and is entering a cache state $B^i_{t_0} \rightarrow B^i_{t_c}$. Directly injecting new context here allows bidirectional attention to handle the context switch where different frames in the cascade are affected with the new context based on their current noise levels. Similar to KV-recaching, our first block is being cached with new context in $B^i_{t_0} \rightarrow B^i_{t_c}$ and provides clean KV features for all future blocks. We find that in practice, videos generated with bidirectional attention over our cascade can blend context much smoother than KV-recaching (more in Supplementary) and are preferred by users (see~\cref{fig:user_study}).

\begin{figure*}[!h]
\centering
\includegraphics[width=\linewidth]{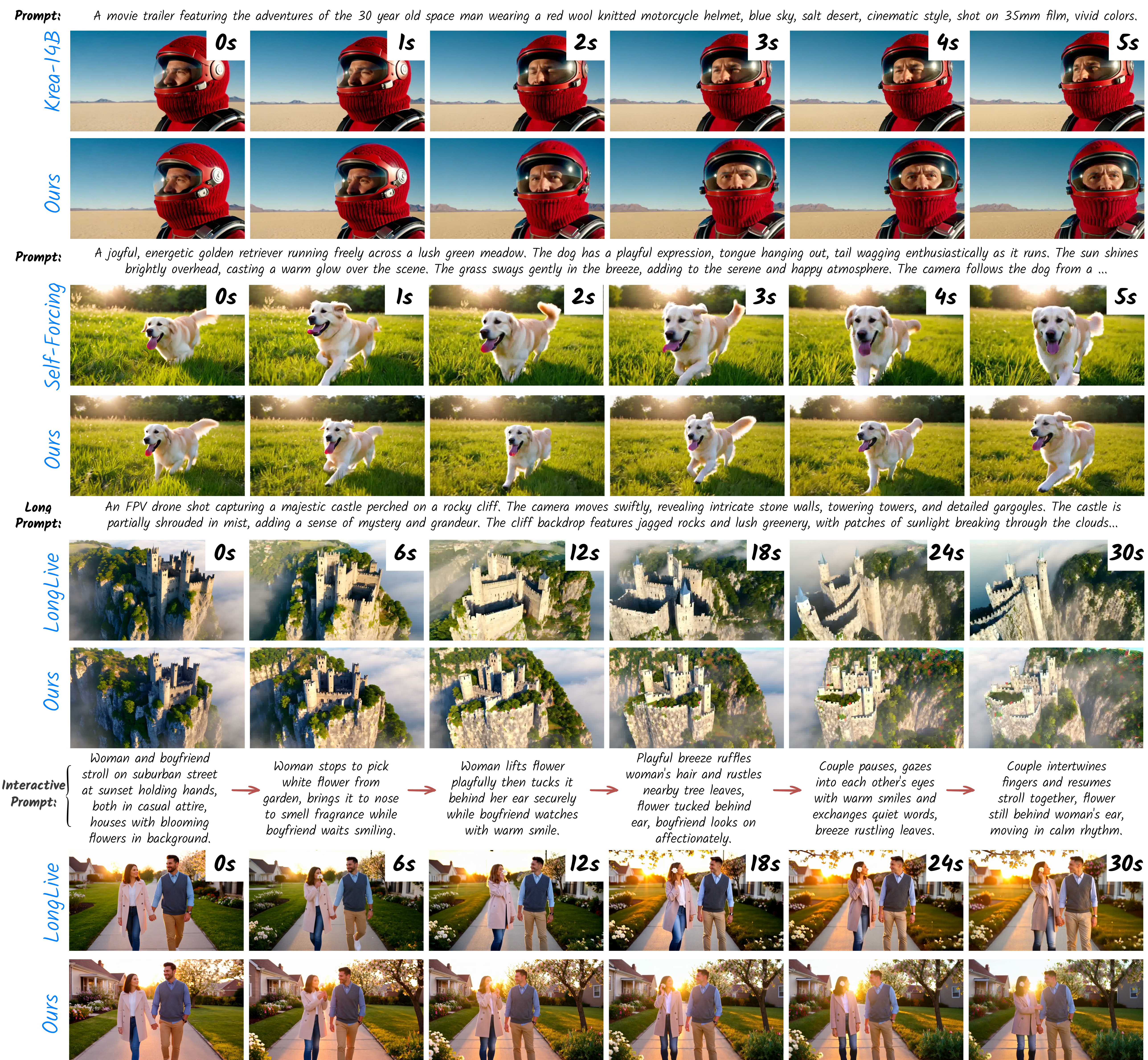}
\vspace{-0.4cm}
\caption{\textbf{Qualitative comparisons: } Comparing our bidirectional inference pipeline with corresponding original inference pipelines}
\vspace{-0.5cm}
\label{fig:qual_comp}
\end{figure*}

\vspace{-0.2cm}
\section{Experiments}

We conduct experiments by adapting causal video models without training. Specifically, we adapt Self-Forcing~\cite{huang2025self}, LongLive~\cite{yang2025longlive} and Krea~\cite{krea_realtime_14b}, obtaining an inference FPS boost of $14.05$, $12.66$ and $8.07$ respectively with multiple GPUs. We choose these particular models for a holistic assessment over different inference domains like short, long, and interactive video generation across different model capacities. For all inference configurations, we use a window size of $7$ blocks, and specifically for LongLive, a sink size of $1$ block where all blocks have $3$ frames each. LongLive's static sink can corrupt dynamic videos, so for interactive generation, we generate our samples without sink while LongLive refreshes sink during KV-recaching. For context change in interactive generation, we simply change the text prompt without incurring any artifacts, which helps us to prevent FPS drop from expensive KV recaching (see~\cref{sec:blockcascading}). We compare quality of videos generated with our pipeline v/s the original inference pipelines with qualitative (\cref{fig:qual_comp}) and quantitative (\cref{sec: quant},~\cref{tab: main}) analysis and a comprehensive user study (\cref{fig:user_study}).

We compare our videos with those generated with (i) \textit{Block-causal video generation pipelines}:  CausVid~\cite{causvid}, Self-Forcing~\cite{huang2025self}, LongLive~\cite{yang2025longlive}, SkyReels-V2~\cite{chen2025skyreels} and MAGI-1~\cite{teng2025magi} and (ii) \textit{Bidirectional Video generation pipelines}: Rolling Forcing~\cite{liu2025rolling}, FastVideo~\cite{zhang2025vsa} and Wan2.1~\cite{wan2025wan}. CausVid~\cite{causvid} learns a block-causal video generative model with the DMD~\cite{yin2024one} objective in~\cref{eq:kldiv}. Self-Forcing~\cite{huang2025self} builds on CausVid, reducing mismatch between training and inference pipelines for block-causal generation by simulating the inference environment during training. LongLive~\cite{yang2025longlive} addresses drifting by training with a permanent attention sink that retains specific context irrespective of causal window size. Separate from these models that are fine-tuned from Wan2.1\cite{wan2025wan}, SkyReels-V2~\cite{chen2025skyreels} and MAGI-1~\cite{teng2025magi} train block-causal pipelines from scratch. Fully bidirectional FastVideo~\cite{zhang2025vsa} distils from Wan2.1~\cite{wan2025wan} using a sparse attention strategy. Rolling-Forcing~\cite{liu2025rolling} implements bidirectional attention on top of block-causal generations strategies and distils from Wan2.1~\cite{wan2025wan} using a rolling bidirectional attention window.


\subsection{FPS Analysis} 

In Block Cascading, FPS initially starts low while blocks are being loaded in the cascade, then stabilises as the cascade is full, and finally goes up towards the end (\cref{fig:abl-multi} and~\cref{fig:abl-single}) as blocks are evicted. In contrast, causal video generation pipelines have initial blocks with smaller attention windows that are faster to generate than later blocks with larger windows. For our experiments, we want to define `Streaming FPS' as the FPS when the attention window is full and new blocks are being generated and decoded with the VAE. This definition reflects the steady-state performance that a user would experience during long, auto-regressive video generation. We report Streaming FPS in~\cref{tab: main} and in~\cref{fig:fps-combined} as the average FPS across 8th and 9th blocks during generation of $13$ blocks with a $7$-block attention window. Additionally, we plot instantaneous FPS observed during generation to illustrate FPS changes (\cref{fig:abl-single} and \cref{fig:abl-multi}) and to highlight FPS drops in interactive video generation ( \cref{fig:fps_recache}).

\begin{figure}
    \centering
    \includegraphics[width=\linewidth]{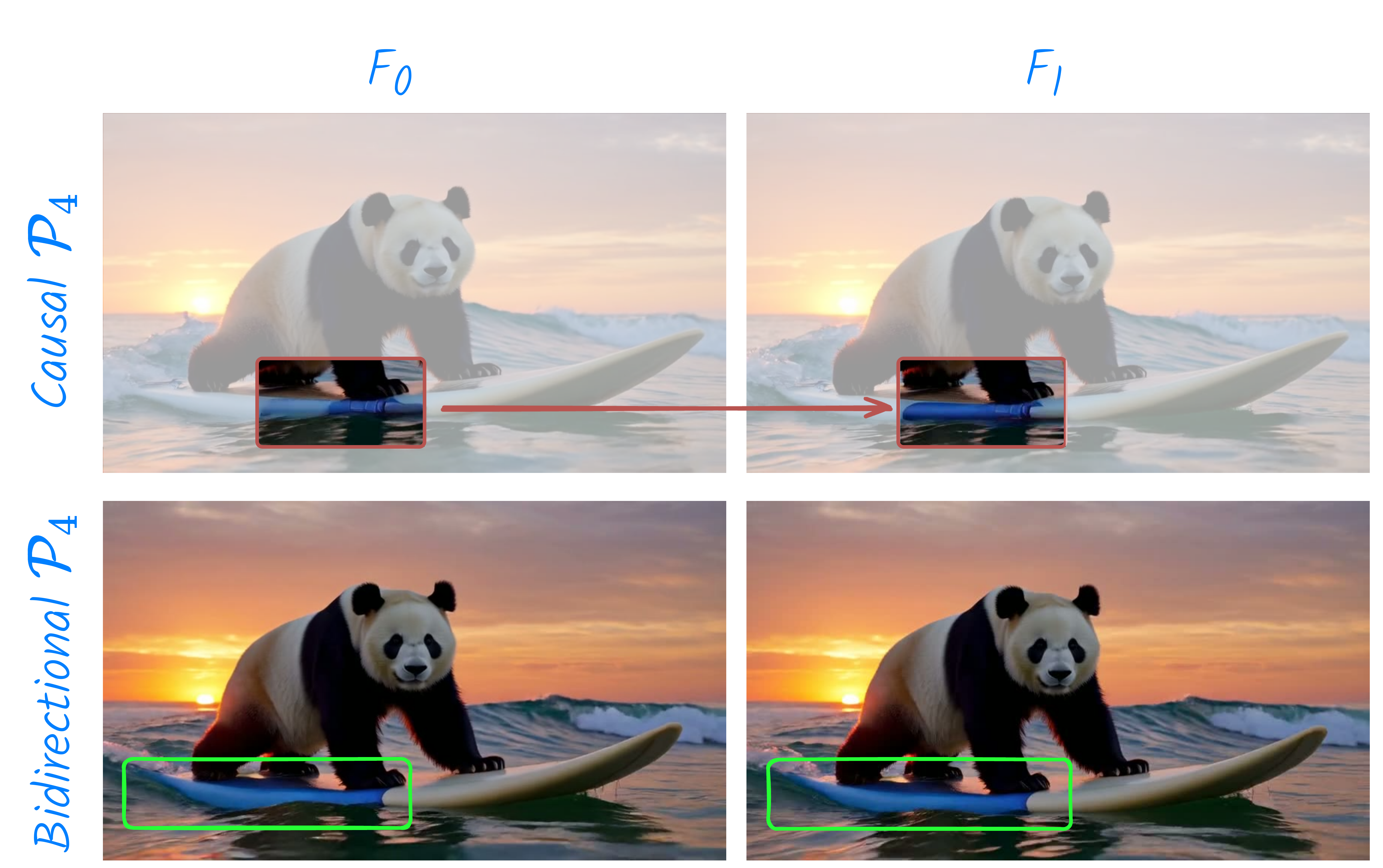}
    \vspace{-0.6cm}
    \caption{\textbf{Bidirectional Inference:} Causal attention in fully parallelised generation  ($\mathcal{P}_4$) can yield artifacts. These can be fixed by using bidirectional attention with same fully parallelised pipeline.}
    \label{fig:causal_artifacts}
    \vspace{-0.3cm}
\end{figure}

\begin{figure}
    \centering
    \includegraphics[width=\linewidth,trim=0 0 100 0, clip]{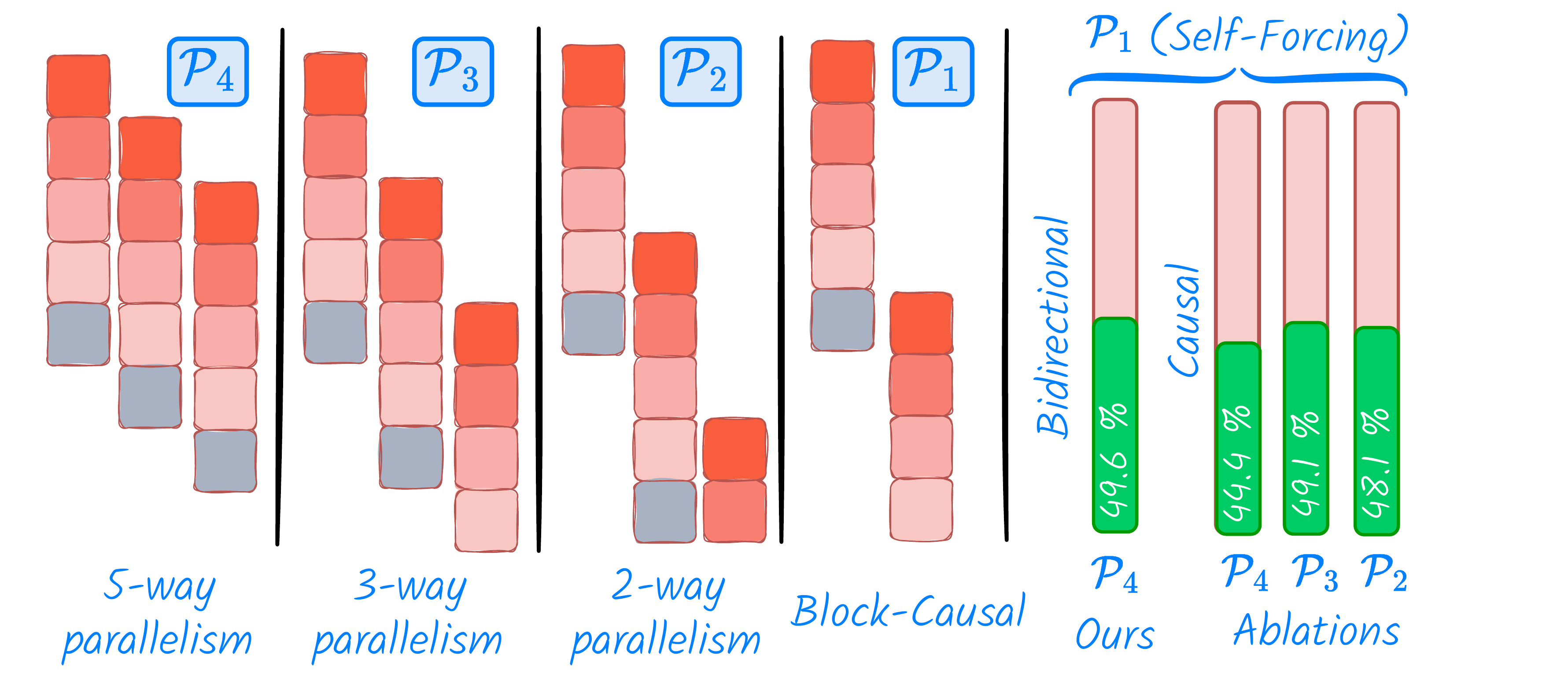}
    \vspace{-0.7cm}
    \caption{\textbf{Cascading Types: } Different types of cascading leads to different levels of parallelism. We conduct a user study to analyse performance across differently cascaded pipelines. At $\mathcal{P}_1$, the pipeline reduces to block-causal pipelines like Self-Forcing~\cite{huang2025self}}
    \label{fig:parallelism}
    \vspace{-0.6cm}
\end{figure}

\subsection{Ablative Studies \label{sec: abl}}
For ablation experiments, we analyse the degree of parallelisation attainable with different forms of Block Cascading, using Self-Forcing~\cite{huang2025self} pre-trained checkpoint as our baseline. We define different parallelization forms based on timesteps for cascading, from $t\in \{t_3,t_2,t_1,t_0,t_c\}$, denoting $\mathcal{P}_i$ for $i$-timestep skip by using $B^j_{t_c-i}$ for denoising $B^{j+1}_{t_0}$. We start with baseline self-forcing as $\mathcal{P}_1$, where blocks are generated sequentially and cached features from $B^i_{t_c}$ are used for denoising $B^{>i}_t$. To reduce dependency on $t_c$, we allow block $B^{i+1}_{t_0}$ to denoise using cache from $B^i_{t_3}$ in $\mathcal{P}_2$ which cascades blocks $B^{i+1}$ and $B^i$. Next, we move block $B^{i+1}_{t_0}$ further up (see~\cref{fig:parallelism}), denoising with block $B^{i}_{t_2}$ for a two block cascade and with $B^{i}_{t_c}$ and $B^{i+2}_{t_0}$ for a three block cascade at $t{=}t_2$. Finally, we use our default configuration $\mathcal{P}_4$, which is a five block cascade (\eg with $\{B^{0}_{t_c}, B^{1}_{t_3}, B^{2}_{t_2}, B^{3}_{t_1}, B^{4}_{t_0}\}$). We observe from~\cref{fig:abl-single} and~\cref{fig:abl-multi} that causal generation from $\mathcal{P}_4$ has the highest FPS both in single (+$2.9$ FPS) and multi GPU (+$14.5$ FPS) environments. However, causal generation can yield artifacts (see~\cref{fig:causal_artifacts}) in early frames with limited clean context. In a single GPU setup, switching to $\mathcal{P}_3$ solves this as cleaner context is available for later frames to reduce artifacts (\cref{fig:parallelism}). In a multi GPU environment, causal and bidirectional attention have marginal difference in FPS, and the best option is to use Bidirectional $\mathcal{P}_4$ that yields high quality with high generation FPS. For analysing video quality with different ablative environments, we perform user studies (see ~\cref{sec: user_study}) observing better performance with bidirectional $\mathcal{P}_4$ and causal $\mathcal{P}_2$ and $\mathcal{P}_3$, compared to causal $\mathcal{P}_4$.

\begin{figure}
    \centering
    \vspace{-0.1cm}
    \includegraphics[width=\linewidth]{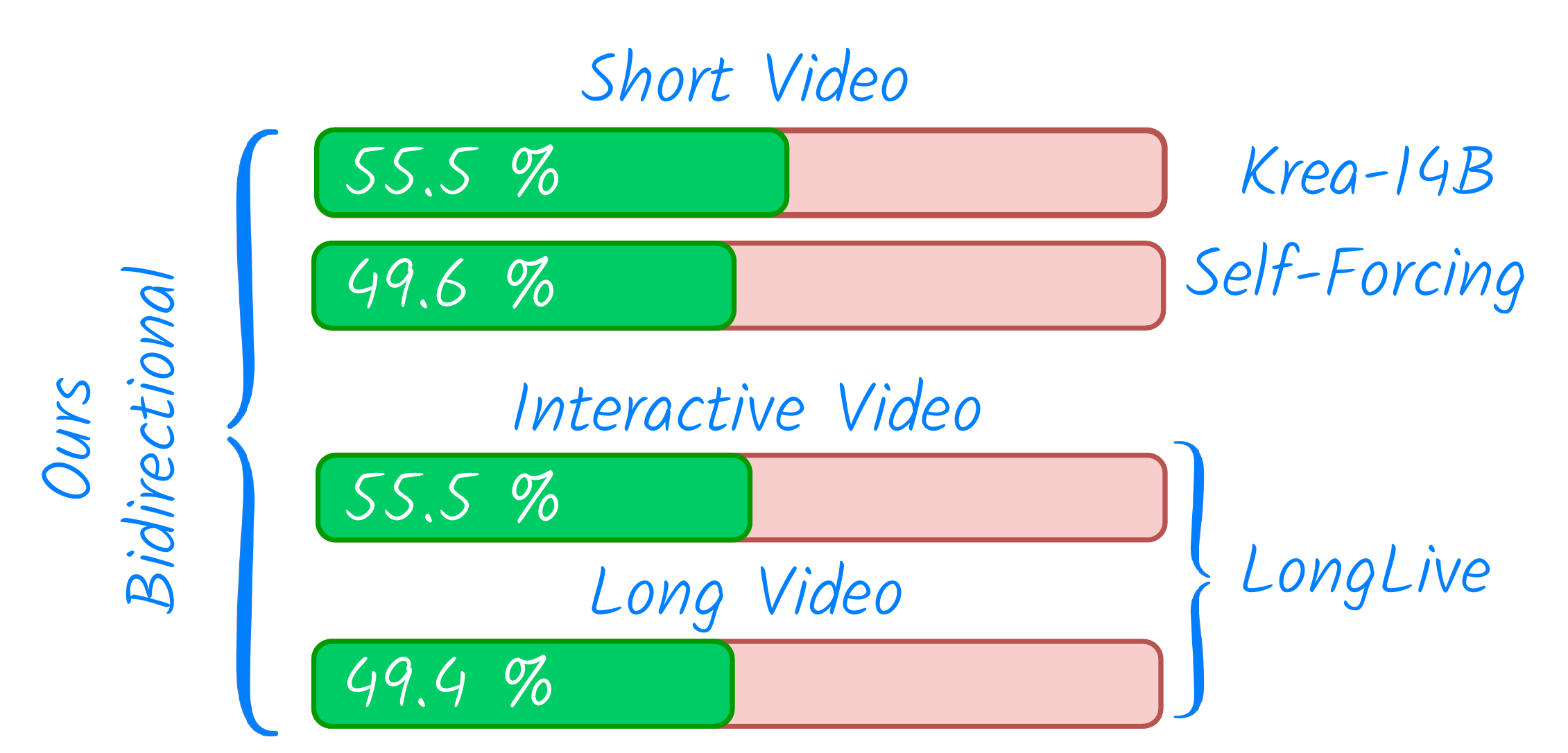}
    \vspace{-0.7cm}
    \caption{\textbf{User Study: } We conduct a user study to analyse video quality from our cascaded bidirectional pipeline against original inference pipelines of other models under different configurations.}
    \label{fig:user_study}
    \vspace{-0.6cm}
\end{figure}

\subsection{User Study \label{sec: user_study}}
To assess degradation in generated videos, we perform user studies where we show users videos generated by Self-Forcing~\cite{huang2025self}, LongLive~\cite{yang2025longlive}, and Krea~\cite{krea_realtime_14b} inference pipelines and those generated by ours. With the same seed and prompt, our method produces videos that closely match those generated by the vanilla inference pipeline (see~\cref{fig:qual_comp}). Therefore, we do not report prompts in our study, except in interactive video analysis where prompts are essential for illustrating context changes. The instruction provided is to choose the video that has higher quality given two competing videos - with each video reviewed by three annotators.
The options available to users are `left', `right', `both' and `none'. We include more details of the user study in supplementary. For Self-Forcing, Krea and our ablative configurations (\cref{sec: abl} and \cref{fig:parallelism}), we generate $418$ short videos from each method with prompts sampled from VBench~\cite{huang2023vbench} extended prompts~\cite{huang2025self}. We evaluate LongLive  on $100$ long videos with prompts sampled from MovieGen~\cite{polyak2024movie} extended prompts and $20$ interactive videos with prompts  from~\cite{yang2025longlive} and Gemini 2.5 Pro~\cite{comanici2025gemini}. Both long videos and interactive videos are generated to a total length of $120$ latent frames ($30$ seconds) while short videos are generated to $21$ latent frames  ($5$ seconds). From results of this user study in~\cref{fig:user_study}, we note that users have a hard time distinguishing our pipeline from original pipeline when compared with Self-Forcing or LongLive. We find our pipeline is preferred in interactive video generation with LongLive and short video generation with Krea, potentially from additional context from bidirectional attention. 

\subsection{Quantitative Analysis}
\label{sec: quant}
We use VBench~\cite{huang2023vbench} for quantitative analysis with prompts re-written using Qwen2.5-7B-Instruct~\cite{Yang2024Qwen25TR} following other methods~\cite{huang2025self}. All samples for evaluation with VBench in~\cref{tab: main} use re-written prompts for video generation. In addition to a brief summary in~\cref{tab: main}, we include a complete table with all VBench scores in Supplementary. From~\cref{tab: main}, inference with our denoising pipeline closely follows scores of the original pipelines without significant drop in quality. During inference, our model is parallelised across 5 GPUs, providing us an avg. boost of $\sim$$2\times$ in FPS.

\begin{table}[!h]
\centering
\vspace{-0.2cm}
\setlength{\tabcolsep}{4.5pt}
\scriptsize
\caption{\textbf{Video quality benchmark:} We report VBench~\cite{huang2023vbench} scores for all competitors. FPS for other methods reported with 1xH100, FPS for ours reported on 5xH100s}
\vspace{-0.3cm}
\label{tab: main}
\begin{tabular}{lcccc}
\toprule
Model & \makecell{Streaming\\FPS ($\uparrow$)} & \makecell{Total\\Score ($\uparrow$)} & \makecell{Quality\\Score ($\uparrow$)} & \makecell{Semantic\\Score ($\uparrow$)} \\
\cmidrule(lr){1-1}\cmidrule(lr){2-2}\cmidrule(lr){3-5}
Wan2.1$^\dagger$~\cite{wan2025wan} & - & 0.8381 & 0.8471 & 0.8019 \\
FastVideo$^\dagger$~\cite{zhang2025vsa} & - & 0.8313 & 0.8647 & 0.6978 \\
\cmidrule(lr){1-1}\cmidrule(lr){2-2}\cmidrule(lr){3-5}
SkyReels-V2~\cite{chen2025skyreels}$^*$ & 0.49 & 0.8267 & 0.8470 & 0.7453 \\
MAGI-1~\cite{teng2025magi}$^*$ & 0.19 & 0.7918 & 0.8204 & 0.6774 \\
CausVid~\cite{causvid} & 16.31 & 0.8502 & 0.8555 & 0.8290 \\
Rolling Forcing~\cite{liu2025rolling} & 17.80 & 0.8287 & 0.8358 & 0.8003 \\
\cmidrule(lr){1-1}\cmidrule(lr){2-2}\cmidrule(lr){3-5}
Self-Forcing~\cite{huang2025self} & 16.31 & 0.8440 & 0.8498 & 0.8206 \\
Ours (w Self-Forcing ckpt) & \textbf{30.36} & 0.8353 & 0.8435 & 0.8024 \\
\cmidrule(lr){1-1}\cmidrule(lr){2-2}\cmidrule(lr){3-5}
LongLive~\cite{yang2025longlive} & 15.60 & 0.8335 & 0.8378 & 0.8162 \\
Ours (w LongLive ckpt) & \textbf{30.46} & 0.8230 & 0.8232 & 0.8223 \\
\cmidrule(lr){1-1}\cmidrule(lr){2-2}\cmidrule(lr){3-5}
Krea-14B~\cite{krea_realtime_14b} & 4.52 & 0.8437 & 0.8471 & 0.8304 \\

Ours (w Krea-14B ckpt) &\textbf{12.59} & 0.8492 & 0.8532 & 0.8332 \\

\bottomrule
\end{tabular}
\vspace{2mm}
\begin{flushleft}
\scriptsize
\vspace{-0.5cm}$^*$Numbers taken from~\cite{huang2025self}, $^\dagger$ Bidirectional pipelines cannot stream.
\vspace{-0.7cm}
\end{flushleft}
\end{table}

\label{sec: exp}
\vspace{-0.2cm}
 \section{Limitations}

We note that window size in pre-training configuration can be a limitation for Block Cascading when full parallelism is desired for optimal FPS boost. 

Empirically, we notice slightly higher drifting in some samples during inference with a $7$-block window, while using a checkpoint that was trained with a $4$-block window \cite{yang2025longlive}. Nevertheless, the degree of parallelism can be reduced ( \eg with $\mathcal{P}_3$ or $\mathcal{P}_2$) for generation with strict small window models at some cost to inference speed. We also note sub-linear scaling with GPUs in multi-GPU environments, making Block Cascading ideal for single video generation but a poor alternative for generating large batches of videos. The latter case is better parallelised with distributed sampling.
\vspace{-0.2cm}
\section{Conclusion}
\label{sec: conclusion}

In conclusion, Block Cascading parallelises inference in block-causal video pipelines without requiring any re-training or fine-tuning. Extensive user studies demonstrate that Block Cascading during inference does not degrade video quality in pre-trained causal pipelines. This allows us to boost streaming inference speed for $1.3$B video models to $30$ FPS and $14$B models to $12.5$ FPS using multiple GPUs, supporting downstream applications like interactive and controllable video generation. Deploying large-scale video models like Krea-14B for dynamic generation and live-viewing bridges the gap between high quality video generation and real-time inference.

{
    \small
    \bibliographystyle{ieeenat_fullname}
    \bibliography{main}
}

\end{document}